\documentclass{article}
\usepackage{ijcai21}

\usepackage{times}
\usepackage{soul}
\usepackage{url}
\usepackage[hidelinks]{hyperref}
\usepackage[utf8]{inputenc}
\usepackage[small]{caption}
\usepackage{graphicx}
\usepackage{amsmath}
\usepackage{amsthm}
\usepackage{booktabs}

\usepackage{fancyhdr,amssymb}
\usepackage{multirow}
\usepackage{tabularx}
\usepackage[ruled,vlined]{algorithm2e}
\usepackage{color,soul}
\usepackage{fancybox}
\usepackage{subcaption}
\usepackage{adjustbox}

\graphicspath{{figs/}}
\newcommand{\RN}[1]{%
	\textup{\lowercase\expandafter{\it \romannumeral#1}}%
}

\ifx\remark\undefined
\newtheorem{remark}{Remark}
\fi

\title{Improve Variational Autoencoder for Text Generation \\ with Discrete Latent Bottleneck}

\author{
Yang Zhao$^1$
\and
Ping Yu$^1$\and
Suchismit Mahapatra$^{2}$\and
Qinliang Su$^3$ \and
Changyou Chen$^1$
\affiliations
$^1$University at Buffalo, $^2$Criteo, $^3$Duke University\\
}

\begin{document}

\maketitle

\begin{abstract}
Variational autoencoders (VAEs) are essential tools in end-to-end representation learning. However, the sequential text generation's common pitfall with VAEs is that the model tends to ignore latent variables with a strong auto-regressive decoder. In this paper, we propose a principled approach to alleviate this issue by applying a discretized bottleneck to enforce an implicit latent feature matching in a more compact latent space. We impose a shared discrete latent space where each input is learned to choose a combination of latent atoms as a regularized latent representation. Our model endows a promising capability to model underlying semantics of discrete sequences and thus provide more interpretative latent structures. Empirically, we demonstrate our model's efficiency and effectiveness on a broad range of tasks, including language modeling, unaligned text style transfer, dialog response generation, and neural machine translation.
\end{abstract}

\section{Introduction}

Auto-encoder models,  which try to learn a function that maps each input to a latent representation and then back to the original data space, are widely used in various NLP tasks such as machine translation \cite{vaswani2017attention} and dialog response generation \cite{olabiyi2019multi}. Variational autoencoders (VAEs)~\cite{kingma2013auto}, by contrast, aim to learn a probabilistic generative model that can generate new instances similar to the original data. Meanwhile, VAEs are endowed with flexible latent representations (\textit{e.g.} style and semantic features) with which one can easily draw diverse and relevant samples from distribution following a decoding scheme. VAEs have achieved tremendous success in generating high-quality images, videos, and speech \cite{van2017neural,razavi2019generating}. At the same time, VAEs have also been applied in NLP to improve traditional maximum-likelihood-estimation (MLE) based models, achieving impressive progress in language modeling \cite{miao2016neural,li2020optimus}, controllable text generation \cite{hu2017toward}, neural machine translation \cite{shah2018generative}, and many other applications.

A well-known pitfall with VAEs, especially in applications of sequence-to-sequence (Seq2Seq) modeling, is a phenomenon called \textit{latent variable collapse} (or posterior collapse) \cite{bowman2015generating}, where an encoder yields meaningless posteriors that collapse to the prior. With this pitfall, VAEs usually fail to learn meaningful representations of individual data samples. Several attempts have been made to alleviate this issue \cite{he2019lagging,fang2019implicit}. However, most of these approaches are heuristic in nature.

Our solution to this problem is motivated by two possible explanations of posterior collapse: $\RN{1})$ Recent research shows that the prior plays a vital role in density estimation \cite{takahashi2019variational}. Although Gaussian prior and posterior are broadly adopted, such simplified priors tend to incur \textit{posterior collapse} for flawed density estimations. To overcome this issue, we argue that \textbf{a flexible prior should be learned simultaneously during training}. 
$\RN{2})$ Related work has also shown that the posterior collapse is caused by a lack of useful latent codes \cite{fu2019cyclical}. Thus, designing \textbf{an effective way of learning good representations without supervision} is the key to address the problem. In this paper, based on the above two arguments, we propose to enforce a discrete latent space for VAEs, namely DB-VAE. The discrete space consists of learnable atoms that are shared by all data inputs. The discrete latent space automatically brings in at least two benefits: $\RN{1})$ The \textbf{discrete atoms are evolving} that are sampled from the prior as the training proceeds instead of being fixed; $\RN{2})$ The discretization performs implicit feature matching and enforces a \textbf{semantically meaningful nearest neighbor structure}.
The contributions of our paper are summarized as follows: 
\begin{itemize}
\setlength\itemsep{-0.3em}
    \item We propose the concept of discretized bottleneck VAEs for RNN-based Seq2Seq models, which can implicitly alleviate the long-standing posterior-collapse issue.
    \item We show how to inject the discretized bottleneck in Seq2Seq models on a variety of NLP tasks. Our DB-VAE can accurately model discrete text without sacrificing reliance on latent representations being employed as a simple plugin. We also find that under our framework, the discrete bottleneck can capture more sentence-level semantic features. 
    \item Inference of the proposed DB-VAE requires a nearest-neighbor (NN) search for the discrete atoms in a latent space. We extend NN to the $k$-NN setting and show that it can provide more reliable translations given one source text in machine translation, which increases the BLEU score. The method is referred to as \textit{top-$k$} search. Similarly, the scheme can also generate diverse responses in the dialog response generation task.
\end{itemize}

\section{Variational Autoencoder}
VAEs consist of two parts, a $\phi$-parameterized encoder (inference network) and a $\theta$-parameterized decoder (generative network). The decoder corresponds to the following generative process for an input $\mathbf{x}$:

\begin{equation}
    \mathbf{z} \sim p(\mathbf{z}), \mathbf{x} \sim p_\theta(\mathbf{x}|\mathbf{z})
\end{equation}

\noindent where $p(\mathbf{z})$ is a pre-defined prior distribution and $p_\theta(\mathbf{x}|\mathbf{z})$ is a conditional distribution induced by a decoder. To learn the parameters $\theta$, one typically maximizes the following marginal log-likelihood: 

\begin{equation}
    \log p_\theta(\mathbf{x}) = \int p(\mathbf{z})  p_\theta(\mathbf{x}|\mathbf{z})d\mathbf{z}
\end{equation} 

Direct optimization of the log-likelihood is usually intractable. VAEs instead parameterize a family of variational distribution $q_\phi(\mathbf{z}|\mathbf{x})$ to approximate the true posterior $p_\theta(\mathbf{z}|\mathbf{x}) \varpropto  p(\mathbf{z})  p_\theta(\mathbf{x}|\mathbf{z})$, ending up optimizing the following evidence lower bound (ELBO): 

\begin{align}\label{eq:elbo}
    \log p_\theta(\mathbf{x}) \geqslant \text{ELBO} &=  \mathbb{E}_{q_\phi(\mathbf{z}|\mathbf{x})} \log p_\theta(\mathbf{x}|\mathbf{z}) \nonumber\\
  & -  \text{KL}(q_\phi(\mathbf{z}|\mathbf{x})||p(\mathbf{z}))
\end{align} 

\noindent where the first term corresponds to the reconstruction error term $\mathcal{L}_{rec}$, and the second term represents the regularizer $\mathcal{L}_{kl}$ with which VAE captures the holistic properties of the input. Most existing works~\cite{li2020optimus} also extends the above ELBO  to a controllable objective $\mathcal{L}$ by introducing a hyper-parameter $\beta$:
\begin{align}\label{eq:control-ojb}
    \mathcal{L} = \mathcal{L}_{rec} + \beta \mathcal{L}_{kl}
\end{align}
Apparently, when we set $\beta=0$, we are learning a vanilla autoencoder. When we let $\beta > 0$, we are learning a smooth latent space.




\section{The Framework}

\subsection{The Model}\label{subsec:seq2seq}
\begin{figure}
    \centering
    \includegraphics[width=0.4\textwidth]{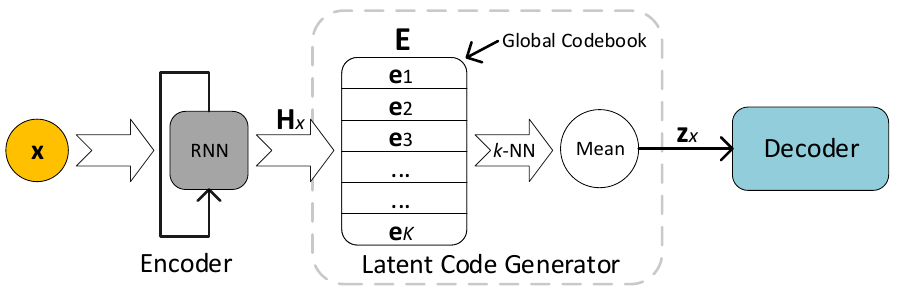}
    \caption{The graphical illustration of the proposed model}
    \label{fig:model}
\end{figure}

We propose to apply a discrete bottleneck to most existing Seq2Seq VAEs. Without loss of generality, we describe our framework under an RNN-based language model setting. As shown in Figure \ref{fig:model}, our model consists of three parts: an encoder, a latent code generator, and a decoder.

\paragraph{Encoder: }
Let an input sequence be defined as  $\mathbf{x}=\left[x_1, ..., x_t, ..., x_T\right]$. The encoder aims at mapping an input token at each time step to a hidden representation. This is implemented by feeding an input sequence to an LSTM encoder, resulting in

\begin{equation}\label{eq:enc}
    \mathbf{h}^{e}_t = \textbf{LSTM}(\mathbf{w}_{\mathbf{x}_{t}}, \mathbf{h}^{e}_{t-1})~,
\end{equation}

\noindent where $\mathbf{w}_{x_t}$ is the word embedding vector of the word $x_t$. The hidden representation of the input sequence $\mathbf{x}$ is $\mathbf{H}_x \triangleq (\mathbf{h}_1^e, \cdots, \mathbf{h}_T^e)$.

\paragraph{Discrete Latent Code Generation: }
Different from the vanilla VAE mechanism, we define a latent code to be a combination of a set of atoms from a global codebook $\mathbf{E} \triangleq [\mathbf{e}_1; \cdots; \mathbf{e}_K] \in \mathbb{R}^{K \times D}$, where $K$ is the codebook size and $D$ is the latent embedding dimension. Our first step is to associate each $\mathbf{h}_t^e \in \mathbf{H}_x$ with one $\mathbf{e}_{\kappa(t)} \in \mathbf{E}$, where $\kappa(t)$ is an index mapping function that maps the index $t$ to another index $k \in [1, \cdots, K]$. Specifically, let $\mathbf{z}_t \triangleq \mathbf{W}_e \mathbf{h}^e_t + \mathbf{b}_e$ with learnable parameters $(\mathbf{W}_e, \mathbf{b}_e)$. We define $\kappa(\cdot)$ to be the nearest neighbor operator:

\begin{equation}\label{eq:enc_linear}
    \kappa(t) \triangleq \arg\min_k  \|\mathbf{z}_t - \mathbf{e}_k\|_2
\end{equation}

Then we define the latent code $\mathbf{z}_x$ of input $\mathbf{x}$ to be an average of multiple samples $\mathbf{z}_t$:

\begin{align}\label{eq:aggregate}
    \mathbf{z}_x \triangleq \textstyle \sum_{t = 1}^T\mathbf{e}_{\kappa(t)} / T~.
\end{align}

\paragraph{Sliced Codebook: }
A common issue of the standard codebook-based vector quantization as proposed in VQ-VAE \cite{van2017neural,razavi2019generating} is codebook collapse which reduces information capacity of the bottleneck, where a large portion of codes are underutilized. This issue is more severe in our text-based applications. To alleviate this issue, we consider using sliced codebook where we consider multiple codebooks~\cite{kaiser2018fast}. Assuming two codebook are used, we first initialize two independent global codebooks, $\mathbf{E}^{(1)}, \mathbf{E}^{(2)} \in \mathbb{R}^{K \times D/2}$. A latent representation $\mathbf{z} \in \mathbb{R}^D$ is then broken up into two slices $\mathbf{z}^{(1)}, \mathbf{z}^{(2)} \in \mathbb{R}^{D/2}$, so that \eqref{eq:enc_linear} can be applied to find respective nearest neighbors $\mathbf{e}^{(1)}$ and $\mathbf{e}^{(2)}$ from $\mathbf{E}^{(1)}$ and $\mathbf{E}^{(2)}$, respectively. The column-wise concatenation $\mathbf{e} \triangleq [\mathbf{e}^{(1)},\mathbf{e}^{(2)}]$ is then taken as the quantized version of $\mathbf{z}$.

\paragraph{Decoder: }
Similar to the encoder, we parameterize the decoder with another LSTM. The target hidden state $\mathbf{h}^{d}_t$ can be progressively calculated as

\begin{equation}
    \mathbf{h}^{d}_t = \textbf{LSTM}([\mathbf{w}_{x_{t}}, \mathbf{z}_x], \mathbf{h}^{d}_{t-1})~,
\end{equation}

Finally, we calculate the output distribution over the entire vocabulary at time $t$ as 

\begin{equation}
    P_t = \text{Softmax}(\mathbf{W}_o \mathbf{h}^{d}_t)
\end{equation}

\subsection{Training}
Learning DB-VAE is divided into two parts: learning the encoder and decoder, and learning the global codebook. 

\paragraph{Learning the Encoder and Decoder: }
The encoder and decoder can be learned by directly optimizing the controlable ELBO as given in~\eqref{eq:control-ojb}.

\paragraph{Learning the Codebook: }
Directly optimizing the codebook with the above $\mathcal{L}_{rec}$ is infeasible because gradients cannot propagate back to the codebook due to the non-differentiable operator defined in \eqref{eq:enc_linear}. To this end, we follow \cite{van2017neural} and define a new objective for updating the codebook. The key observation is that the codebook only appears in \eqref{eq:enc_linear}, thus the goal is to update the codebook such that it makes the distance between a latent code and the corresponding codebook atom minimal. Specifically, the loss is defined as
{\small\begin{align}\label{eq: loss_code}
    \mathcal{L}_{code} = \textstyle \sum_{t=1}^T [\|\text{sg}(\mathbf{z}_t)-\mathbf{e}_{\kappa(t)}\|^2_2 
    + \alpha\|\mathbf{z}_t-\text{sg}(\mathbf{e}_{\kappa(t)})\|^2_2]/T~,
\end{align}}
\noindent where sg($\cdot$) denotes the stop-gradient operator that blocks gradient from flowing through its operand; $\alpha$ is a constant to balance the two terms. The overall learning objective becomes:
\begin{align}
    \mathcal{L} = \mathcal{L}_{rec} + \beta \mathcal{L}_{kl} + \mathcal{L}_{code}
\end{align}
Based on different values of $\beta$, we can derive two versions of DB-VAE:
$(\RN{1})$ $\beta=0$: we replace the continuous latent space with a discrete latent space and directly pass the discrete latent code  $\mathbf{z}_x$  to the decoder.
$(\RN{2})$ $\beta=1$: we only regularize the original VAE with a discrete latent bottleneck without pass the discrete latent code $\mathbf{z}_x$ to the decoder. We name them DB-VAE-$q$ and DB-VAE-$r$ respectively and can treat both of them as regularized AEs where a non-Gaussian prior is imposed~\cite{zhao2017adversarially}. Since DB-VAE-$q$ is more consistent with former applications of vector quantization and different from continuous VAEs, we use DB-VAE-$q$ by default. 

\paragraph{The Overall Algorithm: } 
The full training algorithm is summarized in Algorithm \ref{algo: train_algo}. Both the encoder $f_\phi$ and the decoder $g_\theta$ are implemented as LSTMs. To avoid clutter, $f_\phi$ herein includes the whole procedure \eqref{eq:enc}-\eqref{eq:aggregate}. We find that it is important to balance between learning the encoder-decoder and learning the codebook. At the beginning, if the codebook does not learn as fast as the encoder, there will be a low utilization rate of the codebook, {\it e.g.}, most of the input samples only focus on a limited number of atoms of the codebook. To overcome this issue, we add a straight-through pretraining step, where the decoder is fed with the latent codes directly from the encoder. This ensures that reasonable gradients can be passed through the latent space and the encoder. In the following, we will apply the superscript ``$(i)$'' on a variable (or function) to denote the dependency of the variable to the $i$-th input sample $\tilde{\mathbf{x}}^{(i)}$. To determine whether one should perform a pretraining step, we define a perplexity score $ppl\_code$ to monitor the utilization of the codebook:
\begin{equation}
\begin{split}
    &\mathbf{v} = \frac{1}{mT} \sum_{i=1}^{m}\sum_{t=1}^{T} \text{one\_hot}(\kappa^{(i)}(t)) \\
    &ppl\_code = \exp [-|\mathbf{v} \odot \log(\mathbf{v})|_1]
\end{split}
\label{eq: code_ppl}
\end{equation}
\noindent where $\text{one\_hot}(\kappa^{(i)}(t)$ denotes a all-zero $1 \times K$ vector except the $\kappa^{(i)}(t)$-th bit, which is set to 1. It is clear that the $ppl\_code$ value is large when the elements in $\mathbf{v}$ are close to uniform. Thus it can be used to indicate the utilized rate of the codebook. 

\paragraph{Extension: \textit{top-k} NN search}
In our construction of a latent code, we search the nearest code from the codebook via the index mapping defined in \eqref{eq:enc_linear}. Such a construction endows a limitation where a hidden state from the LSTM only corresponds to one atom from the codebook. This scheme, however, does not fit real applications well. For example, in neural machine translation, one source sentence (one hidden state) can correspond to multiple correct translations (multiple atoms); and in dialog response generation, a good model should be able to generate multiple relevant and diverse responses when same contexts are given. Furthermore, when a VAE is well trained, input texts with similar semantics should be mapped to close clusters in the latent space (see Section~\ref{sec:lm}). As a result, we propose a generalization by extending the 1-NN search to k-NN search when searching the codebook to construct latent codes. In other words, \eqref{eq:enc_linear} returns a set of $k$ indexes, corresponding to the $k$ nearest codebook atoms from the codebook. These atoms are then averaged over the whole sequence to generate the final latent code, as in \eqref{eq:aggregate}. The corresponding algorithm is summarized in Algorithm \ref{algo: ext} in SM~\ref{sec:supplemental}. We provide a case study on machine translation in~\ref{sec:nmt} in SM~\ref{sec:supplemental}.

\paragraph{Comparison with VQ-VAE:}
Our model is closely related to VQ-VAE proposed for image generation \cite{razavi2019generating}. However, we found that directly applying VA-VAE does not work for natural language generation, possibly due to the optimization difficulty in Seq2Seq models, which easily suffers from severe information bottleneck issues. The key differences between our model and VQ-VAE are: $(\RN{1})$ We identify the ability of discretized codebook to prevent posterior collapse; $(\RN{2})$ We pretrain the codebook and use \textit{ppl\_code} to automatically control when to switch to the joint optimization mode; $(\RN{3})$ Inspired by multihead attention mechanism, we have extended the original single codebook learning to sliced codebook learning to avoid codebook collapse in our applications. This is done by concatenating the quantized latent representations extracted from multiple codebooks, which are then learned independently in different spaces. 

\section{Related Work on Posterior Collapse}
Several attempts have been made to alleviate the posterior-collapse issue. Among them, perhaps the simplest solution is via KL cost annealing, where the weight of the KL penalty term is scheduled to increase during training \cite{bowman2015generating} gradually. Later, \cite{fu2019cyclical} proposes a cyclical annealing schedule, which allows progressive learning of more meaningful latent codes by leveraging informative representations of previous cycles as warm re-starts. These approaches tend to encourage the use of latent codes manually, but might hurt a model's density approximation ability as pointed out in \cite{he2019lagging}. Our method differs from these methods in that it maintains a model's representation power while learning an informative latent space by enforcing a discrete bottleneck.

Other solutions include weakening the capacity of a generative network or enhancing the inference network. \cite{yang2017improved} proposes using a dilated CNN as a decoder in VAE by controlling the size of context from previously generated words. \cite{kim2018semi} offers a semi-amortized approach that uses stochastic variational inference to refine an inference network iteratively. This method, however, is expensive to train. Similarly, \cite{he2019lagging} propose a simple yet effective training algorithm that aggressively optimizes the inference network with more updates. Other threads of solutions introduce more complicated priors in the latent space \cite{tomczak2017vae,ziegler2019latent}. \cite{makhzani2015adversarial,joulin2016bag} further replace the KL regularizer with an adversarial regularizer. Our work outperforms these methods without increasing additional training burdens.

In the case of discrete representations in VAE, the most related work is \cite{zhao2018unsupervised}. It applies the Gumbel-Softmax trick \cite{jang2016categorical} to train discrete variables, resulting in effective and interpretable dialog generation. Our approach has broader applicability and is ready to be extended to more NLP tasks. 


\section{Experiments}
We conduct extensive experiments to demonstrate the effectiveness and efficiency of the proposed DB-VAE on various language processing tasks, including language modeling (LM), unaligned text-style transfer, dialog-response generation, and neural machine translation (NMT). Besides, we also evaluate how the codebook size \textit{K} affects a model's performance. The code for reproducing these results will be made publicly available upon acceptance.

\subsection{Language Modeling}\label{sec:lm}
Following \cite{yang2017improved}, we evaluate our model for language modeling, mainly on two large-scale document corpus, \textit{Yahoo}, and \textit{Yelp}. Detailed statistics of the two datasets are given in Table \ref{tab:lm_stat} in the Supplementary Material (SM) \ref{sm:lm}. We first used a simple \textit{synthetic} dataset \cite{he2019lagging} consisting of 16k training sentences and 4k testing sentences to evaluate how the codebook size affects the model's performance.

\paragraph{The Impact of Codebook Size $K$: }
We first investigate the impact of codebook size \textit{K} on the model's behavior.We observe when the codebook size is smaller than $2^{16}$, the bottleneck seems too tight to induce a right amount of representation power. Because validation $ppl$'s are very close when $K \ge 2^{16}$, we adopt $K=2^{16}$ in all our experiments (a trade-off between memory and performance) unless explicitly declared. 

\paragraph{Baseline and Training Details: } Four representative LM models are chosen as baselines, including LSTM-LM, the standard VAE, SA-VAE \cite{kim2018semi}, and Lag-VAE \cite{he2019lagging}, the current state-of-the-art. For fair comparisons, both the inference and generative networks are implemented as a 1-layer LSTM with 1024 hidden units for all models. The word embedding dimension is set to 1024 and the latent size to 32. The SGD optimizer with the same setting is applied to all models. The latent variable is used to initialize the decoder's hidden state and fed as additional input at each time step. We adopt two sliced codebooks since the single codebook fails to work in the experiments.

\paragraph{LM Results: }
The results in terms of perplexity (PPL) and training time are shown in Table \ref{lm_results}. We follow the implementation in \cite{li2020optimus} to report the PPL. As expected, our model achieves the best performance in all the metrics. Remarkably, our model runs almost as fast as the standard VAE. The faster convergence of Lag-VAE initially is because it aggressively trains an encoder, where approximately $50\times$ more data are used to train the encoder in one epoch. We can apply DB-VAE-$r$ for text generation at evaluation.

\begin{table}[!htbp]
    \centering
    \begin{adjustbox}{scale=0.75}
    \begin{tabular}{c|cc|cc}
    \toprule
       \multirow{2}{*}{Models}  & \multicolumn{2}{c|}{{\it Yelp}} & \multicolumn{2}{c}{{\it Yahoo}} \\
        & PPL & Time & PPL & Time \\ \midrule
        LSTM-LM  &  40.64    &      - &     60.75    &    -  \\
        VAE &      40.56     &     5.4 &    61.21   &   6.9 \\
        SA-VAE &    40.39     &     56.3  &  61.59      &   69.2\\
        Lag-VAE  &     37.92     &      20.3 &  56.78   &  15.3 \\ \midrule
        DB-VAE-$r$ (Ours) &    37.41   &    5.4 &   55.87 &   7.0\\
        DB-VAE-$q$ (Ours)  &    \textbf{37.26}   &    5.4 &    \textbf{55.24}  &   7.0 \\
        \bottomrule
    \end{tabular}
    \end{adjustbox}
    \caption{Comparisons on language modeling.}
    \label{lm_results}
\end{table}

\begin{figure}[!htbp]
    \centering
    \includegraphics[width=0.75\linewidth]{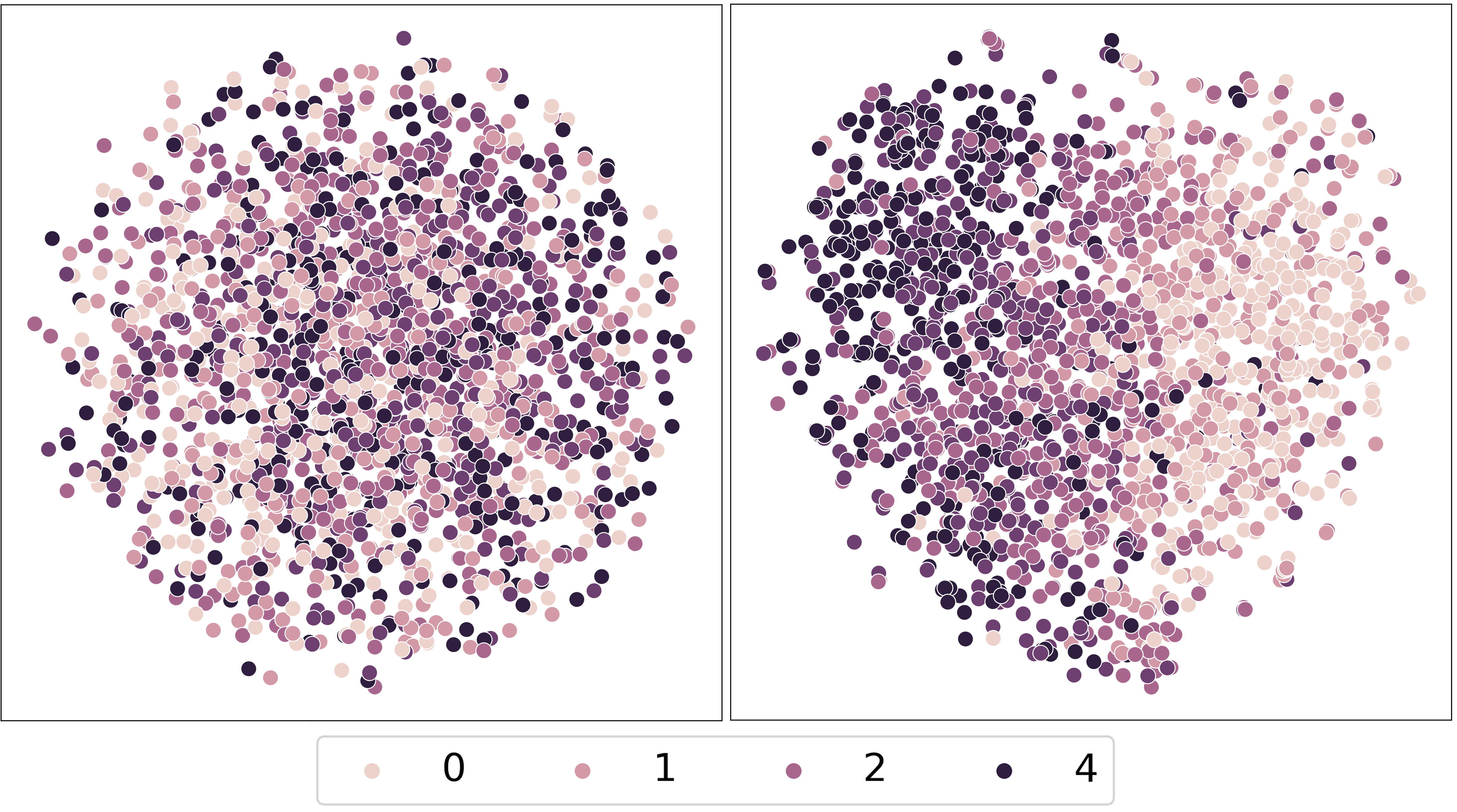}
    \vspace{-2mm}
    \caption{t-SNE embeddings of latent space on \textit{Yelp} corpus. Left: Lag-VAE, Right: DB-VAE. 0-4 represents the review score, from negative to positive.}
    \label{fig:latent_tsne}
    \vspace{-4mm}
\end{figure}

\paragraph{Latent Space Visualization: }
For better understanding, we visualize the latent representations of the whole dataset using t-SNE projection \cite{maaten2008visualizing} in Figure \ref{fig:latent_tsne}. It is seen that our model can learn a much smoother and more separable transition from 0-star to 4-star reviews. To visualize the codebook utilization, we also compute the $\mathbf{v}$ \eqref{eq: code_ppl} on a random batch of testing data after each training epoch. As shown in Figure \ref{fig:code_use} in the SM~\ref{sec:supplemental}, the usage of the codebook becomes more balanced as the training goes on. 

\paragraph{Codebook Interpolation:}\label{para: interpolation}
Particularly in text modeling, when performing a convex combination between any two latent codes $\mathbf{z}_1$ and $\mathbf{z}_2$, the interpolation is equivalent to $\tilde{\mathbf{x}}_\lambda = g_\phi(\lambda \mathbf{z}_1 + (1 - \lambda)\mathbf{z}_2)$. Ideally, adjusting $\lambda$ from 0 to 1 will generate a series of sentences, where $\mathbf{x}_\lambda$ will be less semantically similar with the sentence corresponding to $\mathbf{z}_1$ and much more semantically similar to that of $\mathbf{z}_2$ \cite{berthelot2018understanding}. Table \ref{interpolation_detail} in the SM \ref{sec:supplemental} shows the generated sentences when $\lambda$ ranges from 0.0 to 1.0 with a stepsize of 0.1. Indeed, intermediate sentences $\mathbf{x}_\lambda$ produced by the proposed model can provide a semantically smooth morphing between the two endpoints. 

\paragraph{Comparison with VQ-VAE: } We conduct an ablation study of comparing with original VQ-VAE on the \textit{Yelp} dataset. In both cases of using one codebook and skipping straight-through pretraining, we find that the original VQ-VAE fails to converge where all the data degenerate to only one discrete atom, \textit{e.g.} the utilization of the codebook is similar with the first epoch in Figure~\ref{fig:code_use} in SM. We conjecture that this is because the codebook can't keep up with the model to update the codebook, and the information bottleneck is too tight to perform an effective matching in the latent space.
\begin{table}[!htbp]
\centering
\vspace{-2mm}
\begin{adjustbox}{scale=0.75}
\begin{tabular}{c|c|c|c|c}
\toprule
Model   & Transfer$\uparrow$ & BLEU$\uparrow$  & PPL$\downarrow$ & RPPL$\downarrow$ \\ \midrule
ARAE    & 95.0 & 32.5  & 6.8 & 395  \\
iVAE    & 92.0 & 36.7  & 6.2 & 285  \\  \midrule
DB-VAE & \textbf{97.1} & \textbf{40.2} & \textbf{4.8} & \textbf{254}  \\ \bottomrule
\end{tabular}
\end{adjustbox}
\vspace{-2mm}
\caption{Performances on \textit{Yelp} sentiment transfer}
\label{tab: transfer}
\vspace{-1mm}
\end{table}

\subsection{Unaligned Neural Text Style Transfer}
Next, we evaluate the proposed model's unaligned sentiment transfer task on the \textit{Yelp} dataset. Review ratings above three are considered positive, and those below three are deemed negative. Hence, we split the corpus into two sets of unaligned positive reviews (350k) and negative reviews (250k). The goal of the style transfer task is to change the underlying sentiment between positive and negative reviews.

\paragraph{Experiment Setup: } 
We denote $y$ as the sentiment attribute and construct a decoder to implement the conditional distribution $p(\mathbf{x}|\mathbf{z}, y)$. Following the setup in \cite{zhao2017adversarially,shen2017style}, we train two separate decoders where one is for positive reviews, $p(\mathbf{x}|\mathbf{z}, y=1)$, and the other one is for negative reviews, $p(\mathbf{x}|\mathbf{z}, y=0)$. Normally, the latent prior $p(\mathbf{z})$ will encode all the semantic and attribute information of the input. In the models, we want the attribute information to be excluded from $p(\mathbf{z})$ and let the decoder learn to produce the transferred reviews. According to \cite{zhao2017adversarially}, a classfier $c_\psi$ is introduced to distinguish the latent code's attribute, and adversarially train the encoder to fool the classifier and thus remove the sentiment attribute from the latent space.
\begin{table}[!ht]
\centering
\vspace{-2mm}
\begin{adjustbox}{scale=0.8}
\small
\begin{tabularx}{0.5\textwidth}{lX}
 \multicolumn{2}{l}{\textbf{Negative} $\Rightarrow$ \textbf{Positive}} \\ \toprule
 Input & the staff was very rude as well .\\  
 DB-VAE& the staff here is also fantastic . \\
 ARAE& the staff was very friendly .\\ \midrule
 Input & but , the food is not good .\\  
 DB-VAE & but , the food and brews are the best . \\
 ARAE & well, nice atmosphere with a nice selection .\\ \midrule
 Input & just had a bad experience with a \_num\_ minutes.\\  
 DB-VAE & always a great spot for happy hour or lunch . \\
 ARAE & i love their happy hour .\\ \bottomrule
\\ 
\multicolumn{2}{l}{\textbf{Positive} $\Rightarrow$ \textbf{Negative}} \\\toprule
 Input &  but , it 's worth it !\\
 DB-VAE & however , it 's just ok . \\
 ARAE & but , i was so disappointed .\\ \midrule
Input &  the food is always fresh and tasty .\\
 DB-VAE & the food was n't good , and not fresh . \\
 ARAE & the food was not good but the food was not very good .\\ \midrule
Input &  the service was top notch and so was the food .\\
 DB-VAE & the service was slow and the food was very slow.\\
 ARAE & i was told the server was nice but the food was cold .\\ \bottomrule
\end{tabularx}
\end{adjustbox}
\vspace{-2mm}
\caption{Sentiment transfer results on \textit{Yelp}}
\label{tab: transfer_example}
\end{table}

\paragraph{Baseline: }
We compare our model with two strong baselines: 1) an adversarially regularized autoencoder (ARAE) \cite{zhao2017adversarially}, which learns the prior $p(\mathbf{z})$ via a more expensive and unstable adversarial training; 2) a recently developed implicit deep-latent-variable model (iVAE) \cite{fang2019implicit} that applies sample-based representations of variational distributions. 
\begin{figure*}[t!]
\begin{minipage}[b]{0.68\textwidth} 
\centering
 \includegraphics[height=2.8cm]{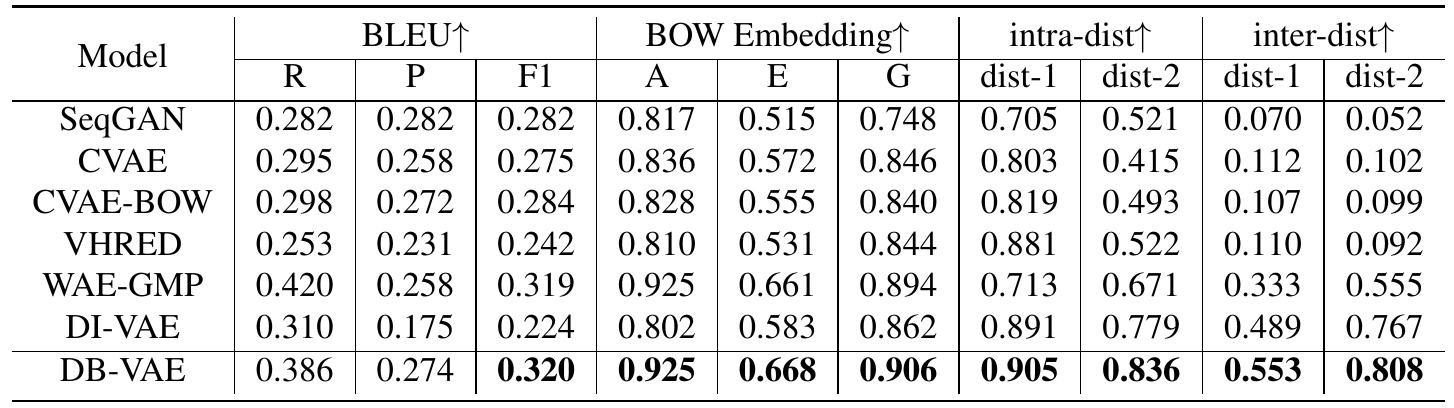}
    \captionof{table}{Comparisons on dialog response generation, \textit{Switchboard Dataset}}
    \label{switchboard}
\end{minipage}
\hspace{-0.1cm}
\begin{minipage}[b]{0.28\textwidth} 
\centering
 \includegraphics[height=1.7cm]{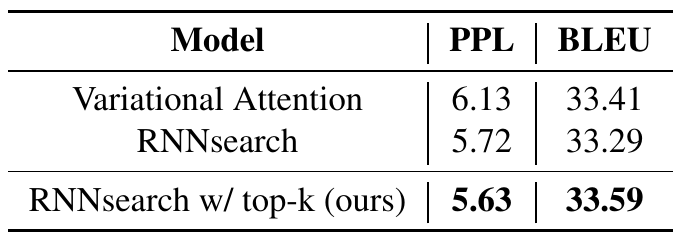}
    \vspace{1.0mm}
    \captionof{table}{Evaluation on NMT}
    \label{tab: bleu}
\end{minipage}
\end{figure*}

\paragraph{Quantitative Metrics: } 
We adopt several quantitative metrics: $(\RN{1})$ Transfer:  it measures the style transfer accuracy evaluated on an automatic classification model (\textit{fastText} library \cite{joulin2016bag}); $(\RN{2})$ BLEU: the consistency between the translated candidate and the original reference; $(\RN{3})$ PPL and Reverse PPL (RPPL): PPL measures the fluency of the generated text, and RPPL works in a reverse fashion, which is computed by training an LM on generated data and evaluated on the original data. The RPPL value may detect mode collapse.

\paragraph{Quantitative Analysis: }
Table \ref{tab: transfer} shows the sentiment transfer results. The proposed method outperforms ARAE in all metrics. On the one hand, in addition to the higher PPL and RPPL, our model preserves the superiority that has already been highlighted in Section \ref{sec:lm}. On the other hand, compared with ARAE, the higher transfer accuracy and BLEU score indicate that our model can capture more sentiment-related information while keeping the original text's grammar structure and the opposite text consistent.

\paragraph{Qualitative Results: }
Some randomly selected examples are give in Table \ref{tab: transfer_example}. We observe that both ARAE and DB-VAE can successfully transfer the sentiment given the input. However, DB-VAE shows better capability in content preserving, and this observation is per the BLEU scores in Table \ref{tab: transfer}.

\subsection{Dialog Response Generation: }
We follow \cite{gu2018dialogwae} and evaluate the proposed model on two widely-used dialog datasets \textit{Switchboard} \cite{godfrey1997switchboard} and \textit{DailyDialog Dataset} \cite{li2017dailydialog}. Responses generated by VAE-based models \cite{zhao2017learning,gu2018dialogwae} are conditioned on the latent variable. This task can examine whether a model can capture a richer latent space and generate more diverse, informative, and consistent responses.

\paragraph{Baselines: }
We compare our model's performance with five representative baselines for dialog modeling:
$(\RN{1})$ SeqGAN: a GAN-based model for sequence generation \cite{yu2017seqgan};
$(\RN{2})$ CVAE: a conditional VAE model \cite{zhao2017learning}; $(\RN{3})$ CVAE-BOW: CVAE with bag-of-word loss \cite{zhao2017learning}; $(\RN{4})$ VHRED: a hierarchical VAE model \cite{serban2017hierarchical}; $(\RN{5})$ WAE-GMP: a conditional Wasserstein autoencoder with a Gaussian mixture prior network \cite{gu2018dialogwae}, which holds the state-of-the-art. $(\RN{6})$ DI-VAE: a discrete VAE which is most related to our work.

\paragraph{Quantitative Metrics: } Follow the evaluation setup in \cite{gu2018dialogwae}, three evaluation metrics (see details in \ref{sm:dialog}) are used: \\
$(\RN{1})$ Sentence-level BLEU, which works by counting n-grams in the candidate (generated) sentences to n-grams in the reference text.
$(\RN{2})$ BOW Embedding, which calculates the cosine similarity of bag-of-words embedding between the candidate and the reference. 
$(\RN{3})$ Distinct, which computes the diversity of the generated responses.

\paragraph{Quantitative Analysis: }
Table \ref{switchboard} and Table \ref{dailydialog} show the quantitative results of our model and other strong baselines on \textit{Switchboard} and \textit{DailyDialog}. Our model outperforms the baselines in most metrics. Although our method obtains a similar BLEU score as WAE-GMP, the inter-dist and intra-dist scores are much higher. In terms of intra-dist, the dist-1 and dist-2 on \textit{Switchboard} are 19.2\% and 24.6\% higher than WAE-GMP. This indicates that our model is capable of generating less repeated n-grams in each response. The inter-dist, dist-1 and dist-2 are even 66.1\% and 45.6\% higher than WAE-GMP, meaning that our model generates much more diverse responses than WAE-GMP. 


\section{Conclusion}
We propose the DB-VAE, a variant of VAE that uses a discretized bottleneck obtained from a global codebook for latent representations. The proposed DB-VAE can provide the right balance between optimizing the inference network and the generative network. Moreover, our DB-VAE can also interpret richer semantic information of discrete structured sequences. Extensive experiments demonstrate the improved effectiveness of the proposed approach compared. Future works include extending discrete bottleneck to large-scale NLP models such as the Transformer and BERT.

\bibliographystyle{named}
\bibliography{ijcai21}

\newpage
\clearpage
\appendix

\section{Supplemental Material}\label{sec:supplemental}

\subsection{Language modeling}\label{sm:lm}

\paragraph{LM Datasets} We follow the original dataset split in \cite{yang2017improved,he2019lagging} where 100k are sampled as
training and 10k as validation and test from the respective
partitions. Dataset details are given in Table~\ref{tab:lm_stat}.
\begin{table}[!htbp]
\centering
\vspace{-2mm}
\begin{adjustbox}{scale=0.9}
\begin{tabular}{c|c|c|c} \toprule
Corpus & \#vocabulary & \#sentences &  avg.length\\ \midrule
Yahoo & 20001 &  10000 & 80   \\
Yelp & 19997 &  10000 &  97  \\ \bottomrule
\end{tabular}
\end{adjustbox}
\caption{Statistics of LM datasets}
\label{tab:lm_stat}
\end{table}

\paragraph{Training Setting}
We follow the training setting in~\cite{he2019lagging}. The SGD optimizer starts with a learining rate 1.0 and is decayed by a factor of 2 if the validation loss has not improved in 2 epochs and terminate training once the
learning rate has decayed a total of 5 times. The LSTM
parameters are initialized from $\mathcal{U}(-0.1, 0.1)$, and embedding parameters are initialized from $\mathcal{U}(-0.01, 0.01)$. A dropout of 0.5 is applied in the decoder for both the embedding and the output.

\begin{figure}[!htbp]
\centering
\includegraphics[width=0.4\textwidth]{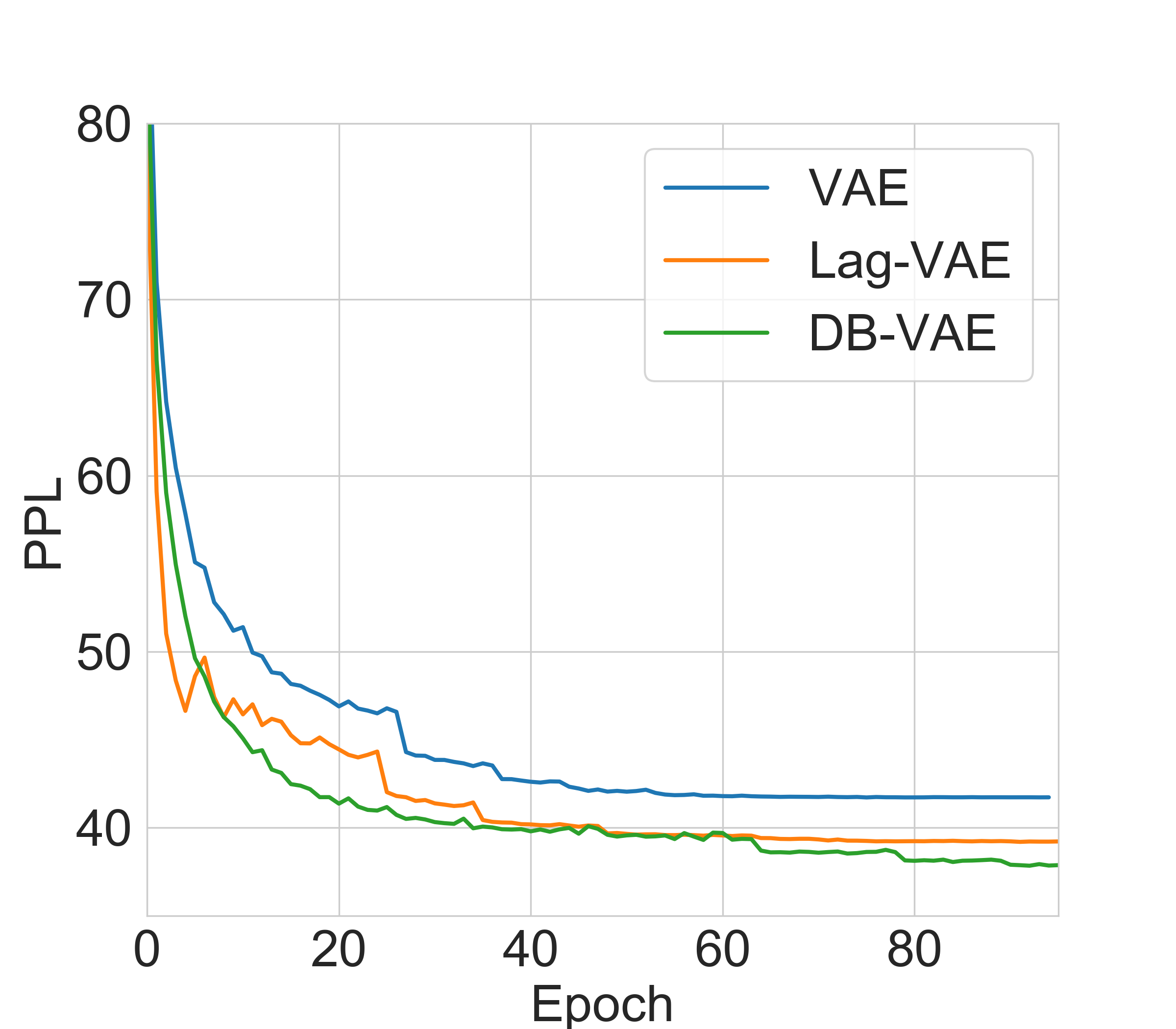}
\caption{Learning curves of VAE, Lag-VAE and DB-VAE-$q$ on \textit{Yelp}}
\label{fig: lm_recon_sm}
\end{figure}

\paragraph{More LM Results: } Figure~\ref{fig: lm_recon_sm} shows the learning curve on \textit{Yelp} testing dataset. Consistently, we observe that the proposed DB-VAE converge faster than the vanilla VAE and Lag-VAE to a lower perplexity.

\begin{figure}[!htbp]
    \centering
    \includegraphics[width=0.8\linewidth]{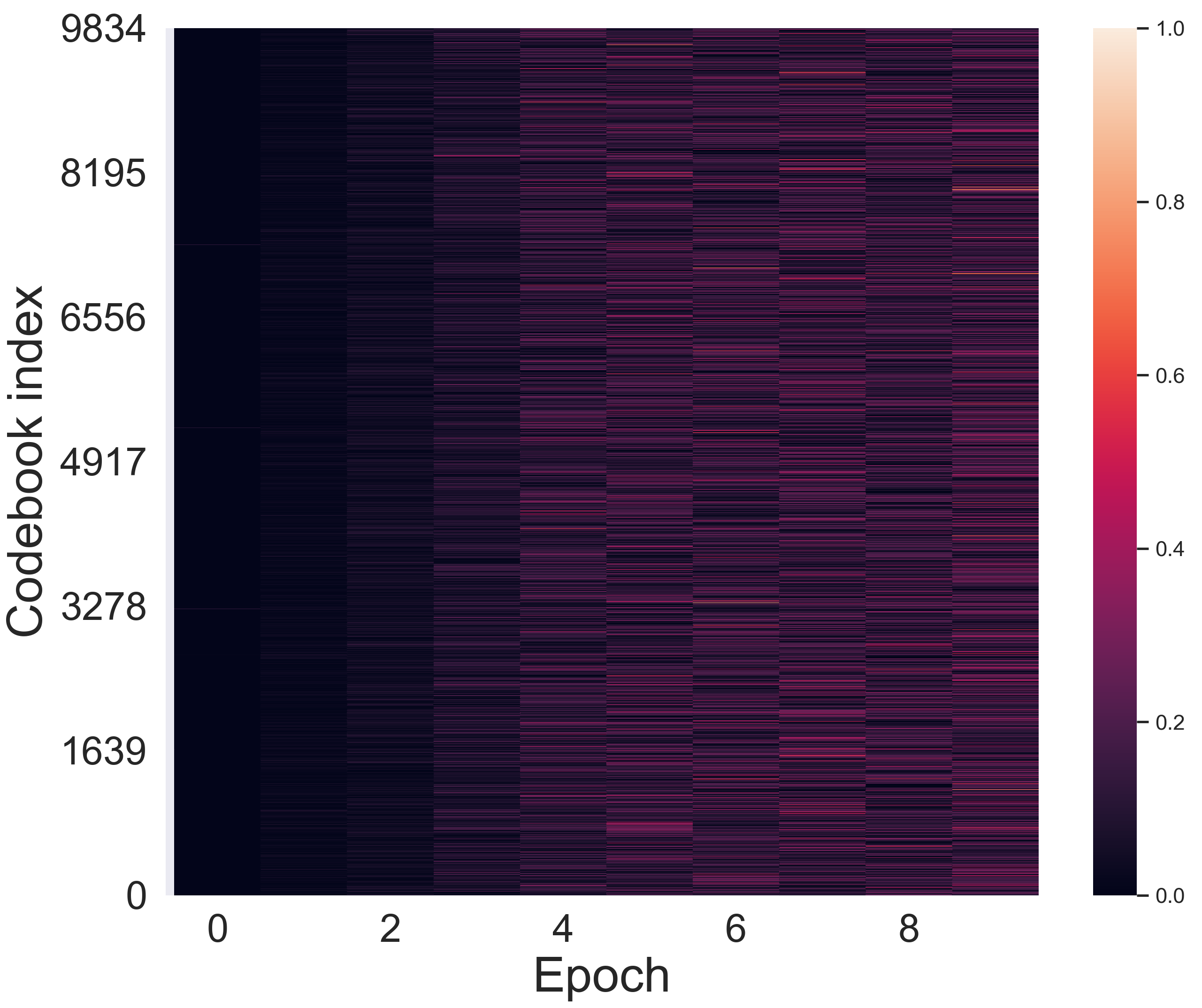}

    \caption{The heatmap of codebook learning on \textit{Yelp}. The \textit{x}-axis corresponds to the training epoch, and \textit{y}-axis corresponds to indices of different codes.}
    \label{fig:code_use}

\end{figure}

\paragraph{Codebook Learning}
Figure~\ref{fig:code_use} shows the evolving utilization of one codebook.

\begin{figure}[!htbp]
\centering
\includegraphics[width=0.8\linewidth]{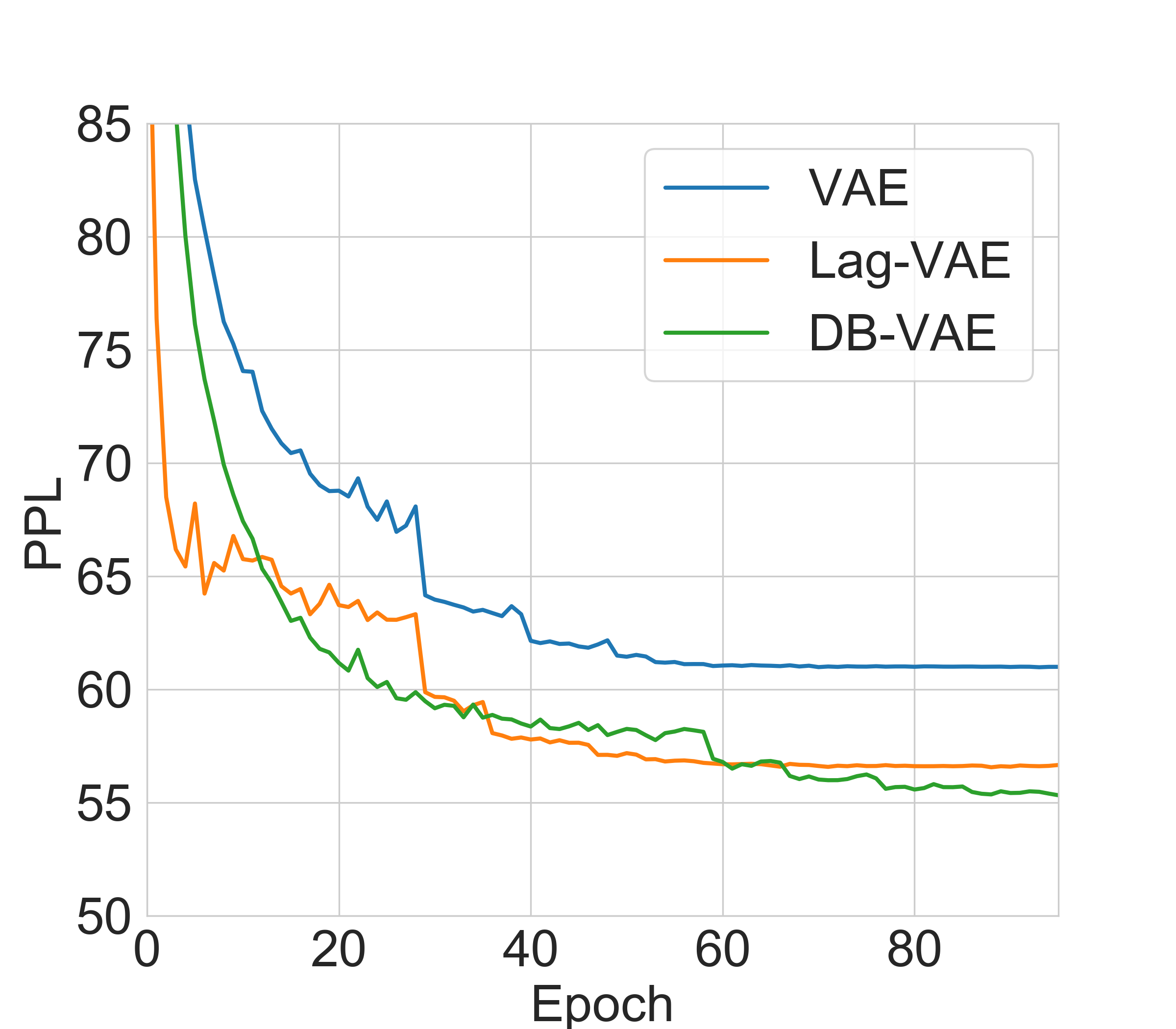}
\caption{Learning curves of VAE, Lag-VAE and DB-VAE-$q$ on \textit{Yahoo}.}
\label{fig: lm_recon}
\end{figure}

\begin{figure}[!htbp]
    \centering
    \includegraphics[width=0.8\linewidth]{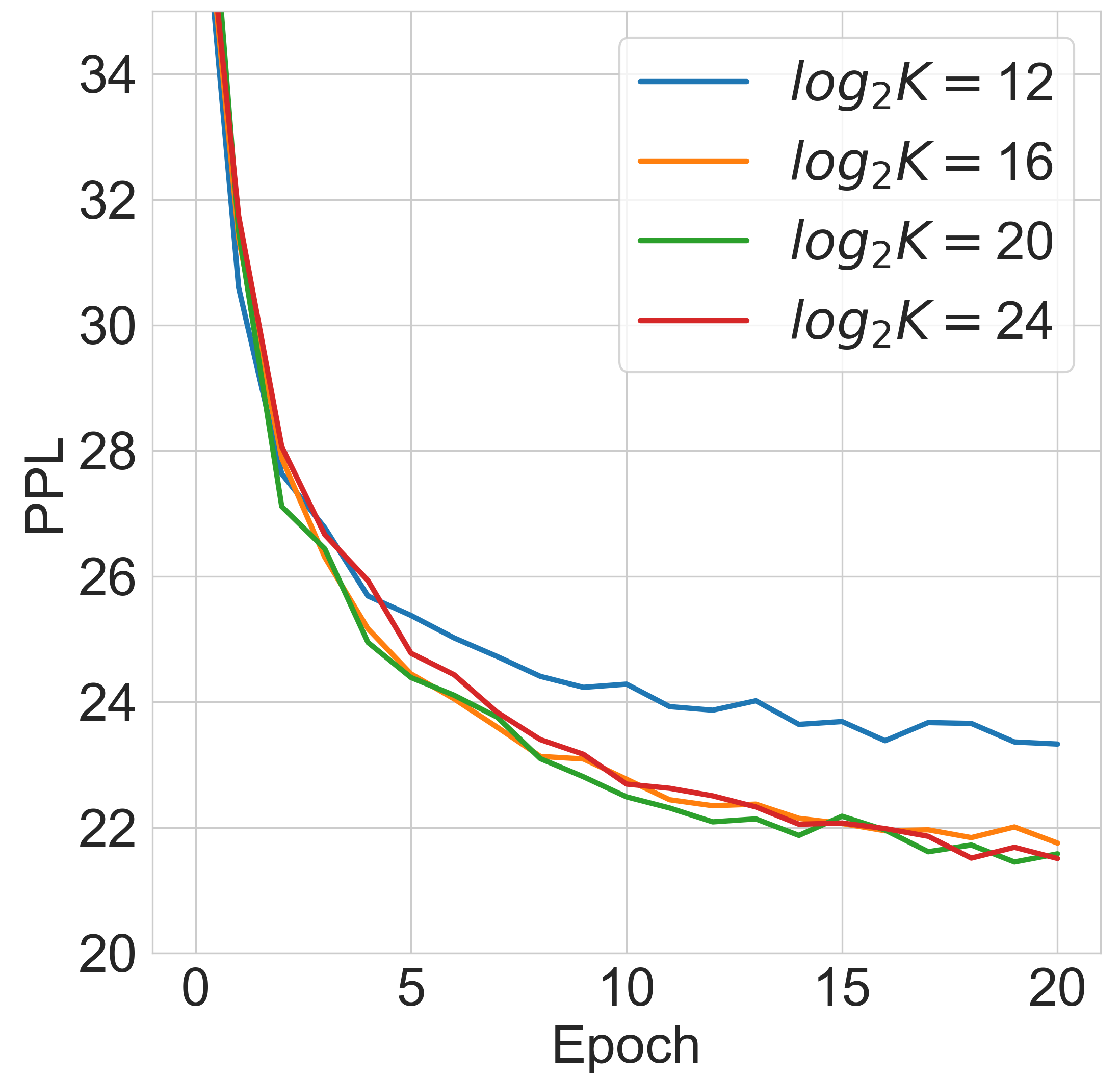}
    \caption{Learning curves with differnt codebook size \textit{K}}
    \label{fig:impact_K}
\end{figure}
 
\begin{algorithm}[htbp!]
\small
\textbf{Require}: encoder $f_\phi$, decoder $g_\theta$, codebook $\mathbf{E}$, threshold $\sigma$ and batch size $m$\\
 \textbf{Step 1}: Straight-through pretraining \\
  \While{$ppl\_code$ $\le \sigma$  }{
  Sample $\{\mathbf{x}^{(i)}\}_{i=1}^{m} \sim q(\mathbf{x})$ \\ 
  Compute $\mathbf{z}_x^{(i)}=f_\phi(\mathbf{x}^{(i)}), \tilde{\mathbf{x}}^{(i)}=g_{\theta}(\mathbf{z}_x^{(i)})$\\
  Optimize the $\mathcal{L}_{rec}+\beta \mathcal{L}_{kl}$ to train $f_\phi, g_\theta$\\
  Optimize $\mathcal{L}_{code}$ to learn $\mathbf{E}$
 }
 \textbf{Step 2}: Joint training \\
 \While{done}{
  Sample $\{\mathbf{x}^{(i)}\}_{i=1}^{m} \sim q(\mathbf{x})$ \\ 
  Compute $\mathbf{z}_x^{(i)}, \tilde{\mathbf{x}}^{(i)}=g_{\theta}(\mathbf{z}_x^{(i)})$\\
  Optimize $\mathcal{L}_{code}$ to learn $\mathbf{E}$\\
  Backprop $\mathcal{L}$ to train $f_\phi, g_\theta$\\
 }
 \caption{\small{DB-VAE training}}
 \label{algo: train_algo}
\end{algorithm}

\paragraph{\textit{top-k} NN Search}
The \textit{top-k} NN search algorithm is given as follows. We can effortlessly apply it to diverse dialog response generation and neural machine translation.
\begin{algorithm}[ht!]
\small
\KwResult{$\{\tilde{\mathbf{x}}^{(i)}\}_{i=1}^k$}
\textbf{Require:}encoder $f_\theta$, decoder $g_\phi$, codebook $\mathbf{E}$\\
\While{done}{
 Sample $\mathbf{x} \sim q(\mathbf{x})$ \\
 Find k-NN instead of 1-NN as in Eq.\eqref{eq:enc_linear} to calculate $\{\mathbf{z}_x^{(i)}\}_{i=1}^k$ \\
 Generate $\{\tilde{\mathbf{x}}^{(i)}=g_\theta(\mathbf{z}_x^{(i)})\}_{i=1}^k$
 }
 \caption{\textit{top-k} NN Search Extension}
 \label{algo: ext}
\end{algorithm}

\begin{figure*}
    \centering
    \includegraphics[width=0.95\textwidth]{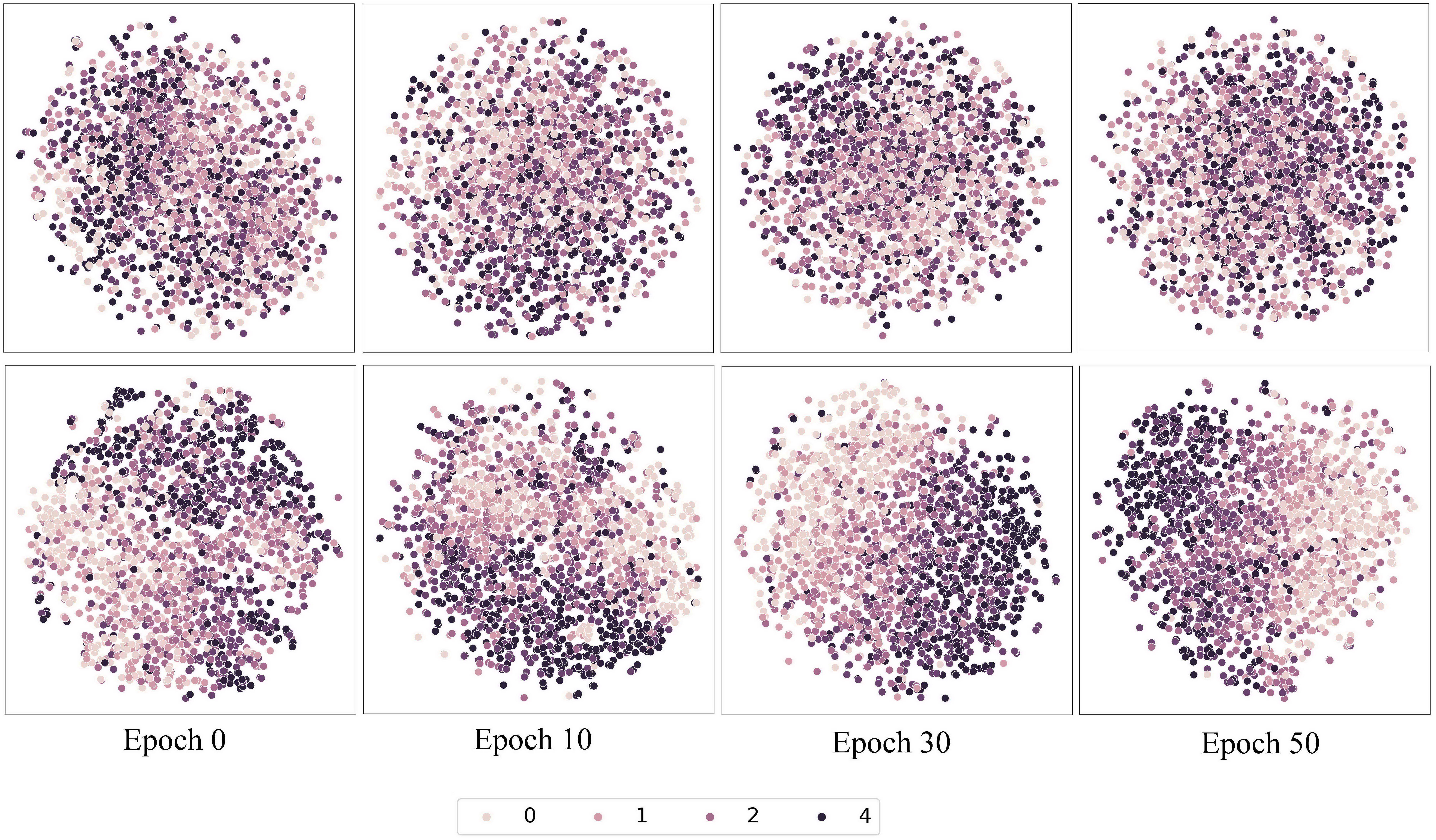}
    \caption{t-SNE embeddings of latent space on \textit{Yelp} corpus. Top: Lag-VAE, Bottom: DB-VAE. 0-4 represents the review score, from negative to positive.}
    \label{fig:latent_tsne_all}
\end{figure*}

\begin{table*}[!htbp]
\begin{tabularx}{\textwidth}{l|X} \toprule
$\lambda$ & \textbf{Generated intermediate sentences} \\ \midrule
0.0 & had a great experience at this place ! i had a great experience with the staff and the staff was very friendly and helpful ! i had a great experience and i will definitely be back ! \\
0.1 & had a great experience here ! the staff was very friendly and helpful ! i had a great time and i will definitely be back ! \\
0.2 & stopped in for a quick bite before heading out to the airport . i had the chicken and waffles and it was delicious ! i would definitely recommend this place to anyone looking for a great breakfast ! \\
0.3 & stopped in for a quick bite before heading out to the airport . i had the chicken and waffles and it was delicious ! the service was fast and friendly . i will definitely be back ! \\
0.4 & stopped in for a quick bite before heading out to the airport . i had the chicken and waffles and it was delicious ! the service was friendly and fast . i 'll be back !\\
0.5 &  my husband and i stopped in for a quick bite before heading out to the airport . we were seated right away and we were seated right away . our server was very friendly and helpful . the food was pretty good and the service was great . \\
0.6 & my husband and i stopped in for a quick bite before heading out to the airport . we were seated right \\
0.7 & this was my first time here and i will definitely be back . the service was fast and friendly and the food was delicious . i 'll be back .\\
0.8 & this was my first time here and i will definitely be back . the service was good , the food was good , and the prices were reasonable . i 'll be back .  \\
0.9 & this place was pretty good . i had the chicken and waffles and it was pretty good . i 'd definitely go back . \\
1.0 & this place was pretty good . i had the pulled pork sandwich and it was pretty good , but nothing special . the fries were pretty good though .\\ \bottomrule
\end{tabularx}
\caption{Detailed interpolation results}
\label{interpolation_detail}
\end{table*}

\begin{table*}[!htbp]
\centering
\begin{tabular}{c|c|c|c|c|c|c|c|c|c|c}
\toprule
\multirow{2}{*}{Model} & \multicolumn{3}{c|}{BLEU$\uparrow$} & \multicolumn{3}{c|}{BOW Embedding$\uparrow$} & \multicolumn{2}{c|}{intra-dist$\uparrow$} & \multicolumn{2}{c}{inter-dist$\uparrow$}\\ \cline{2-11} 
    & R  & P  & F1  & A  & E   & G  & dist-1  & dist-2  & dist-1  & dist-2 \\ 
    \hline
    SeqGAN & 0.270 & 0.270  &  0.270   & 0.907 & 0.495  &   0.774   &   0.747   &  0.806 &    0.075   &    0.081  \\
    CVAE&  0.265  &   0.222  &   0.242   &  0.923 &  0.543  &  0.811  &  0.938     &   0.973   &   0.177 & 0.222 \\ 
    CVAE-BOW &  0.256      & 0.224 & 0.239 &   0.923 &  0.540 &  0.812 &  0.949  &  0.976 & 0.165 & 0.206  \\
    VHRED &  0.271 &  0.260 & 0.265 & 0.892  & 0.507  &  0.786  &  0.633  & 0.711  &   0.071   &    0.089     \\ 
    WAE-GMP&  0.372 & 0.286 &  0.323 & 0.952 &   0.591 &  0.853  &   0.754 & 0.892 & 0.313 &    0.597  \\
    DI-VAE &  0.323 &  0.190 & 0.239 & 0.874  & 0.600  & 0.814 & 0.947 & 0.963 & 0.500 &  0.718    \\ \hline
    DB-VAE  & \textbf{0.373} &  0.276 & 0.317 & 0.944  & \textbf{0.615} & 0.839  & \textbf{0.954}  & \textbf{0.997} & 0.467 & \textbf{0.787} \\
    \bottomrule
\end{tabular}
\caption{Performance comparison on dialog response generation, \textit{DailyDialog Dataset}}
\label{dailydialog}
\end{table*}

\subsection{Dialog response generation}\label{sm:dialog}
Detailed evaluation metrics used in dialog-response-generation task: \\
$(\RN{1})$ Sentence-level BLEU, which works by counting n-grams in the candidate (generated) sentences to n-grams in the reference text. To compute the score, the setting is identical  WAE-GMP \cite{gu2018dialogwae} where 10 responses (candidates) are sampled from the models for each test context. $K$ is set to 10 in Algorithm \ref{algo: ext}. The precision and recall of BLEU are defined in \cite{zhao2017learning}.\\
$(\RN{2})$ BOW Embedding, which calculates the cosine similarity of bag-of-words embedding between the candidate and the reference. We adopt three metrics here to compute the similarity, \textit{greedy} \cite{rus2012comparison}, \textit{average} \cite{mitchell2008vector} and \textit{extreme} \cite{forgues2014bootstrapping}.\\
$(\RN{3})$ Distinct, which computes the diversity of the generated responses. dist-n is defined as the ratio of unique n-grams (n=1,2) over all n-grams in the generated responses. As multiple responses are sampled from the models, we can define intra-dist as the average of distinct values within each sampled response and inter-dist as the distinct value among all sampled responses. 

We adopt the experimental setting in~\cite{gu2018dialogwae} and simply plug in the proposed discrete bottleneck.

\subsection{Extension: RNN-based NMT Model}\label{sec:nmt}
We finally evaluate our model with the proposed \textit{top-k} NN search on the German-English translation task. Our model is built on a baseline RNNsearch architecture \cite{bahdanau2014neural}. The recently proposed variational attention model \cite{deng2018latent} is also adopted as a baseline.

\begin{figure}[!htbp]
    \centering
    \includegraphics[width=0.95\linewidth]{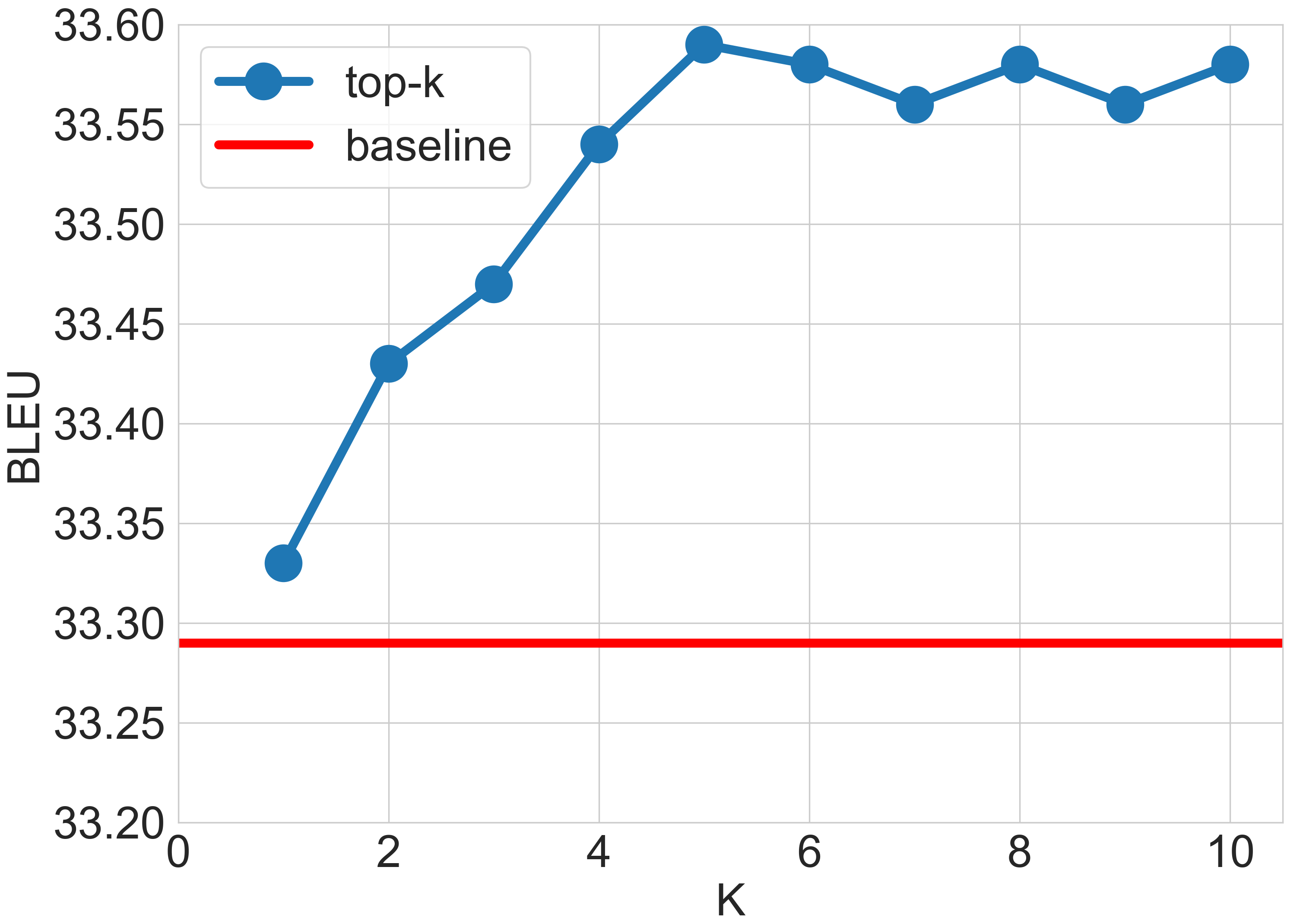}
    \caption{BLEU score on IWLST14}
    \label{fig:bleu}
\end{figure}

We use the IWLST14 dataset \cite{cettolo2014report}, which is a standard benchmark for experimental NMT models. Dataset and practical settings are given in SM~\ref{sec:nmt}.
Results averaged by 5 different runs are reported in Table \ref{tab: bleu} and in Figure \ref{fig:bleu} (in SM). Note the attention mechanism is used in RNNsearch, where each progressed state in the decoder side has direct access to the state on the encoder side. Although we only discretize the encoder's final hidden state as formulated in Section \ref{subsec:seq2seq}, a notable improvement on the PPL and BLEU score is still observed. Following Algorithm \ref{algo: ext}, as we increase the value of $k$ from 1 to 10, the BLEU score continues increasing until $k$ reaches 5. The reason might be that the top-5 latent codes have already encoded most source-target combinations. Besides, the BLEU score is as low as 26.1 when we choose the farthest latent code from the codebook instead. These validate the effectiveness of our proposed top-k inference strategy, which applies to most RNN-based autoencoder models.

 IWLST14 dataset contains around 153K, 7K and 7K sentences for training, validation and testing, respectively. The same preprocessing as in \cite{ott2018scaling} is applied. As for the architecture, both the encoder and the decoder have one layer, each with 512-dimensional embedding. For BLEU evaluation, the beam size in beam search is 5. The library \textit{Fairseq} \cite{ott2019fairseq} is adopted as the codebase. The codebook size $K$ is set to $2^{20}$, and only the final hidden state of the encoder passes through the discretized bottleneck. 
\end{document}